\documentclass{fcs}
\usepackage{bm}
\usepackage{multirow}
\usepackage{enumerate}
\usepackage{wrapfig}
\usepackage{graphicx}
\usepackage{hyperref}
\usepackage{subfigure}
\usepackage{amssymb}
\usepackage{pifont}
\usepackage{algorithm, algorithmic}
\usepackage[misc,geometry]{ifsym}
\volumn{,2024,18(6):186604}
\doi{:10.1007/s11704-023-2791-8}
\articletype{RESEARCH~ARTICLE}
\copynote{{\copyright} Higher Education Press 2024}
\ratime{Received Dec 24, 2023; accepted Jul 6, 2023; issued Sep 28,2023; published Dec 15, 2024. This paper will appear in Front. Comput. Sci.}
\email{$shang@nwpu.edu.cn$}
\title{Federated Learning-Outcome Prediction\\ with Multi-layer Privacy Protection}
\author{Yupei ZHANG$^{1,2}$, Yuxin LI$^{1,2}$, Yifei WANG$^{1,2}$, Shuangshuang WEI$^{1,2}$, Yunan XU$^{1,2}$, Xuequn SHANG(\Letter)$^{1,2}$ }
\address{{1\quad School of Computer Science, Northwestern Polytechnical University, Xi'an 710129, China}\\
{2\quad MIIT Lab of Big Data Storage and Management, Xi'an 710129, China}}

\begin{document}
\maketitle
\setcounter{page}{1}
\setlength{\baselineskip}{14pt}

\begin{abstract}
 Learning-outcome prediction (LOP) is a long-standing and critical problem in educational routes. Many studies have contributed to developing effective models while often suffering from data shortage and weak generalization to various institutions due to the privacy-protection issue. To this end, this study proposes a distributed grade prediction model, dubbed FecMap, by exploiting the federated learning (FL) framework that preserves the private data of local clients and communicates with others through a global generalized model. FecMap considers local subspace learning (LSL), which explicitly learns the local features against the global features, and multi-layer privacy protection (MPP), which hierarchically protects the private features, including model-shareable features and not-allowably shared features, to achieve client-specific classifiers of high performance on LOP per institution. FecMap is then achieved in an iteration manner with all datasets distributed on clients by training a local neural network composed of a global part, a local part, and a classification head in clients and averaging the global parts from clients on the server. To evaluate the FecMap model, we collected three higher-educational datasets of student academic records from engineering majors. Experiment results manifest that FecMap benefits from the proposed LSL and MPP and achieves steady performance on the task of LOP, compared with the state-of-the-art models. This study makes a fresh attempt at the use of federated learning in the learning-analytical task, potentially paving the way to facilitating personalized education with privacy protection.
\end{abstract}

\Keywords{federated learning, local subspace learning, hierarchical privacy protection, learning outcome prediction, privacy-protected representation learning}

\section{Introduction}

\noindent Learning-outcome prediction (LOP) is based on students' learning record data to predict their learning performance in future courses, which helps to achieve personalized instructional resource allocation. Due to its fundamental position in learning analytics and educational data mining, many studies have been developed by using various statistical and machine learning models in wide situations \cite{zhang2021predicting}. There are many kinds of noise, such as slipping, guessing, and subjectively grading \cite{zhang2020meta}, so LOP is often formulated into grade classification. 
\begin{figure}[h]
\centering
\includegraphics[width=0.48\textwidth]{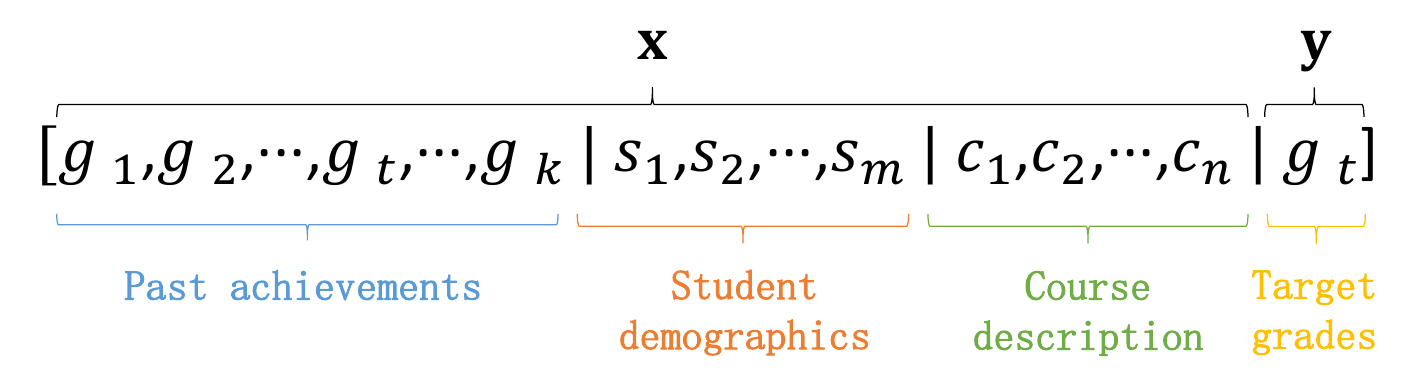}
\caption{The formulated data point for a student. Past achievements refer to the student's grades in other courses, where ${g_\mathit{i}}$ is the grade in the $i$-th course. Student demographics and course description refers to relevant information about students and courses, where ${s_\mathit{i}}$ and ${c_\mathit{i}}$ are the features, and ${g_\mathit{t}}$ is the target grade to be predicted.}
\label{grade}
\end{figure}

As is shown in Fig. \ref{grade}, this study takes each grade record corresponding to the event that a student enrolled in a course and achieved the grade. In mathematics, let ${g_\mathit{i}}$, ${s_\mathit{i}}$, ${c_\mathit{i}}$ and ${g_\mathit{t}}$ be the grade obtained at the $i$-th course, the student’s $\mathit{i}$-th feature, the $\mathit{i}$-th course’s feature and the target grade related to this record, respectively. Here, we cast the data feature vector into $\mathbf{x}=(\mathbf{g}_i;\mathbf{s_\mathit{i}};\mathbf{c_\mathit{i}})$ where $\mathbf{g}_i=[g_1,g_2,\cdots,g_k]$, $\mathbf{s}_i=[s_1,s_2,\cdots,s_m]$ and $\mathbf{c}_i=[c_1,c_2,\cdots,c_n]$. And then we denote a grade record by $(\mathbf{x},\mathit{y})$ where $\mathit{y}=\mathbf{g_\mathit{t}}$. With these notations, the LOP problem can be defined as follows. Given $n$ data points $\left\{\mathbf{x}_i, y_i\right\}_{i=1}^n$, LOP aims to construct a mapping $\mathit{y}=\mathcal{F}(\mathbf{x})$ which uses students' past data $\mathbf{x}_t$ to predict the likelihood grade of the student on the next enrolled course, i.e., $\mathcal{F}(\mathbf{x}_\mathit{t})=\mathit{y}_\mathit{t}$. Towards LOP, the existing machine learning research methods are mainly divided into matrix factorization-based methods \cite{symeonidis2019multi, zhang2020graphs}, similarity-based methods \cite{bydvzovska2015student}, and mapping-based methods \cite{al2017student, polyzou2019feature}. Refer to the review paper of Zhang et al. \cite{zhang2021educational} for more details. 

Nevertheless, data shortage and low generalization caused by data privacy issues often limit current methods. The privacy issue is often ignored, but important to protect personal information and sensitive data in schools, leading to data isolation. Thus, the educational data at hand is always a small sample set with a distribution bias concerning the whole picture of educational behaviors. LOP models on such data often have low generalization accuracy for other schools.

Federated learning (FL) has the potential to handle the above data-isolated problem. FL enables many clients jointly train a machine-learning model while keeping their local data decentralized. Usually, FL is trained in an iterative manner: In each round of communication, users train client models $\mathcal{M}_{\mathbf{W}^i}$ on their respective datasets and then upload them to the server for computation, and the server calculates the global model $\mathcal{G}_{\mathbf{W}^*}$ and then distributes it to all client models. More specifically, given multiple datasets for $m$ subtasks or clients, the $m$ client-model parameters are denoted by
\begin{equation}
    \mathit{W}=\left\{\mathbf{W}^1,\mathbf{W}^2,\cdots,\mathbf{W}^m\right\}
\end{equation}
Then, the general problem of federated learning (FL) with $m$ clients aims to 
\begin{equation} \label{myeq2}
\arg \min _{\mathbf{W}^*, W} \sum_{t=1}^m \mathcal{G}_{\mathbf{W}^*}\left(\sum_{i=1}^{n_t} \mathcal{L}_t\left(\mathcal{M}_{\mathbf{W}^t}\left(\mathbf{x}_i\right)\right)+\lambda \mathcal{E}\left(\mathbf{W}^t\right)\right)
\end{equation}
where $\mathcal{G}_{\mathbf{W}^*}(\cdot)$ indicates the global model with parameter ${\mathbf{W}^*}$, $n_t$ is the number of samples in the $t$-th dataset, $\mathcal{L}_t(\cdot)$ is the loss function of the task indicated by $t$, $\mathcal{E}$ is a regularization imposed on the model parameters $\mathbf{W}^t$, $\lambda$ is the trade-off parameter between loss and its regularization and $\mathcal{M}(\cdot)$ is the local model. Many studies are witnessed in this emerging field, mainly divided into privacy-protection FL framework \cite{li2021ditto}, the federated machine learning models \cite{li2020federated2,collins2021exploiting}, and the convergence of FL algorithm \cite{haddadpour2019convergence}. More FL models can be found in the review paper \cite{li2020federated}.

However, the existing FL models often focus on acquiring a better global model while failing to maintain the whole local representation in clients. Moreover, there is no FL work to keep private information in a hierarchy that is commonly encountered in real-world applications, such as finance and healthcare. In education, hierarchical privacy can be public, shareable after encryption, and not allowable. Hence, we proposed a novel FL framework with local subspace learning (LSL) and multi-layer privacy protection (MPP), dubbed FecMap, to achieve client-specific classifiers of high performance on LOP in each institution. 
\\\\
% \subsection{Overview}
1.1 \quad Overview

\begin{figure}[t]
\centering
\includegraphics[width=0.5\textwidth]{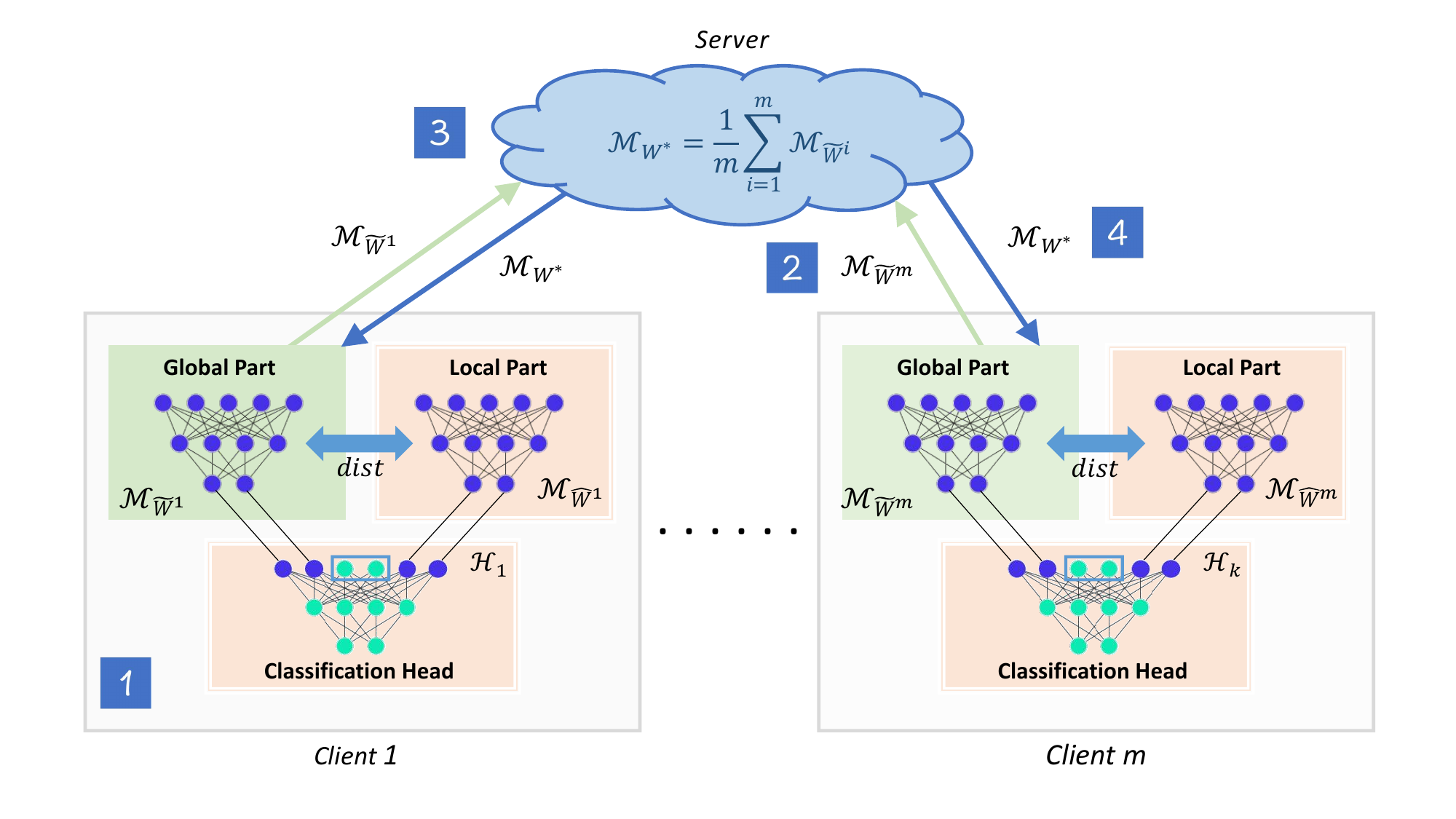}
\caption{The FecMap model trained in an iterative manner. An FL communication is completed by (1) training the local model, (2) uploading to the server, (3) computing the global model, and (4) updating the local model.}
\label{Framework}
\end{figure}

\noindent Figure \ref{Framework} illustrates the proposed workflow of FecMap running on $m$ clients. The workflow is composed of four steps as follows: \ding{172} training client model on the privacy-protected data, where FecMap learns a global part, a local part, and a classification head in each client; \ding{173} uploading the global part of client model, where the local part and the classification head are kept locally; \ding{174} computing the global model in the server, which is achieved by weighted averaging over all the global parts from clients; \ding{175} returning the global model to update the global part of the client model. As usual, FecMap learns the model parameters in the iteration manner over the four steps. Our contributions are majorly three-fold.

\begin{itemize}
\item LSL divides the feature space of the local client into a local subspace and a global subspace. By introducing the maximization subspace separation criterion, we obtained a shared representation for FL communication and a private representation for local classification.
\item MPP divides the local client's privacy features into a multi-level privacy hierarchy. By partitioning the model parameters to learn model-shareable features and not-allowed features, the FL global communication and the local private protection are achieved, respectively.
\item A FL framework, i.e., FecMap, is proposed for LOP, which is carried out online. FecMap trains a local deep network consisting of a global part, a local part, and a classification head. On the server, model averaging is performed on the global parts from the clients.
\end{itemize}

The remainder of the paper is organized as follows. We review the previous studies on FL in Section 2. The object function and the FecMap framework are presented in Section 3. Section 4 shows the experimental results on three datasets collected from our institution. Section 5 concludes this paper.

\section{Related work}
Current FL studies mainly focus on data distribution, federated machine-learning models, privacy-protected strategy, and personalization \cite{li2021survey, tan2022towards, li2021personalized}. i) The FL studies of data distribution are to handle the three data shapes in practice, i.e., horizontal FL where all clients have identical features but different samples \cite{mcmahan2017communication}, vertical FL where all clients have identical samples but different features, and hybrid FL where all clients have small overlaps on both samples and features \cite{tan2022towards}. ii) The studies on federated machine-learning models are to extend traditional models to train in the FL schema, e.g., federated ridge regression \cite{CHEN201834}, federated K-means \cite{dennis2021heterogeneity}, federated principal component analysis \cite{ribero2022federating}, federated online learning \cite{zhou2019privacy}, etc. iii) The studies of privacy-protected strategy are to control the parameter communication for holding privacy, such as the differential-privacy strategy \cite{ribero2022federating}, the gradient-based aggregation \cite{CHEN201834}, and the homomorphic encryption \cite{li2021survey}. iv) The studies on personalization are to deal with the heterogeneity of features and samples \cite{zhang2023doubly, ma2022layer, li2021fedphp, zhang2022personalized}. Smith et al. proposed a model of federated multi-task learning, i.e., the MOCHA model, by integrating the association between clients \cite{smith2017federated}. PFedLA trains a dedicated hyper-network per client in the server to implement personalized FL \cite{ma2022layer}. Li et al. proposed Ditto for personalization by balancing the global model and local models and training a local classification head \cite{li2021ditto}. The work \cite{zhang2022personalized} achieves personalized federated learning through variational Bayesian inference. In addition, many FL applications have already been conducted on training recommendation systems on mobile devices \cite{mcmahan2017communication, duan2020self}, predicting treatment outcomes in hospitals \cite{bercea2022federated}, and managing health in home \cite{wu2020fedhome, wang2022feco}.

In recent years, extensive FL models have been proposed to pursue better performance. To generate high-quality models for clients, many studies are concatenated on handling the heterogeneity of local datasets, such as the multiple-center FL \cite{long2023multi} and clustered FL \cite{sattler2020clustered}, and enhancing local models, such as the FedRep \cite{collins2021exploiting} and FedBN \cite{li2021fedbn}. On the other hand, personalized FL models attract much attention by balancing the server-client dissimilarity \cite{li2021ditto, zhang2022personalized}, inheriting private models \cite{li2021fedphp}, maximizing correlation with sparse and hierarchical extensions \cite{li2021personalized} and integrating the similarity between clients \cite{ZHANG2022109960, 10.1145/3558005}. In addition, the system designation of FL is another important direction, such as the scaled FL \cite{bonawitz2019towards} and the communication-efficient FL \cite{mcmahan2017communication}.

However, these current FL methods fail to maintain the specific features of local clients that can help improve the local model performance. Moreover, there is no FL work to keep private information in a hierarchy. The recent work \cite{li2021aggregate} privatizes some layers in DNN-based federated learning under non-i.i.d. settings to facilitate the learning process. Motivated by aggregation and privatization, we designed a new federated framework by introducing local subspace learning (LSL) and multi-layer privacy protection (MPP) to boost personalization in federated learning.

\section{The Proposed Method}
This section presents our proposed FecMap shown in Figure \ref{Framework}. This paper uses italic lowercase letters for variables (e.g., $\mathit{x}$), bold lowercase letters for vectors (e.g., $\mathbf{x}$), bold capital letters for matrices (e.g., $\mathbf{X}$), italic capital letters for sets (e.g., $\mathit{X}$), and calligraphic letters for functions (e.g., $\mathcal{X}$).
\\\\
% \subsection{Motivation}
3.1 \quad Motivation
\\\\
% \subsubsection{Local Subspace Learning (LSL)}
3.1.1 \quad Local Subspace Learning (LSL)

\noindent In each round of FL communication, users train client models $\mathcal{M}_{\mathbf{W}^i}$ on their respective datasets and then upload them to the server to obtain the weight of the global model $\mathcal{G}_{\mathbf{W}^*}$ by
\begin{equation}
\mathbf{W}^*=\sum_{i=1}^m p_i  \mathbf{W}^i
\end{equation}
where $p_i$ is the weight assigned to the client model $\mathcal{M}_{\mathbf{W}^i}$ based on local data. Usually, $p_i=1/m$. However, there is a model bias between $\mathcal{M}_{\mathbf{W}^i}$ and $\mathcal{G}_{\mathbf{W}^*}$, which can be defined as
\begin{equation}
    \widehat{\mathbf{W}}^i=\operatorname{dist}\left(\mathbf{W}^*, \mathbf{W}^i\right)
\end{equation}
where $\operatorname{dist}\left(\mathbf{A},\mathbf{B}\right)$ is used to measure the difference between $\mathbf{A}$ and $\mathbf{B}$. This bias makes it often difficult for the global model to be the best model for clients.

Specifically, federated learning's global model primarily captures common subspaces and may ignore individualized features. Existing FL models mostly incentivize the global model $\mathcal{G}_{\mathbf{W}^*}$ to fit shared features, yet fail to take into account the bias of the global model to local datasets. A strategy is to set another mapping function, i.e., the local submodel $\mathcal{M}_{\widehat{\mathbf{W}}^i}$, to learn $\operatorname{dist}\left(\cdot\right)$ and $\mathbf{W}^i$.
% \subsubsection{Multi-layer Privacy Protection (MPP)}
\\\\
3.1.2 \quad Multi-layer Privacy Protection (MPP)

\noindent Existing FL studies encrypt all model parameters to upload to the server but fail to consider the levels of privacy security. Inspired by this, an intuitional idea is to protect privacy features by dividing them into shareable features $\mathbf{F}^*$ and not-allowed features $\mathbf{F}^{'}$. Wherein, shareable features $\mathbf{F}^*$ refers to features in the local model that can be shared with the server. This part of the feature does not have strong privacy and can be uploaded to the server in traditional federated communication; not-allowed features $\mathbf{F}^{'}$ refer to features that are trained only in the local model and do not participate in the communication of the server.

In this study, student grades $\mathbf{g}_i=[g_1,g_2,\cdots,g_k]$ are shareable information that we use to capture the global features by communicating with other clients. Student personal information $\mathbf{s_\mathit{i}}=[s_1,s_2,\cdots,s_m]$ is a sensitive not-allowed feature that we use to complement the local features. As the course description may involve sensitive student information, we also consider $\mathbf{c_\mathit{i}}=[c_1,c_2,\cdots,c_n]$ as the not-allowed feature.
\\\\
% \subsection{Model objective}
3.2 \quad Model objective

\noindent \textit{The proposed FecMap combines the strategy for LSL}. Let $\mathcal{M}_{\widetilde{\mathbf{W}}^m}$ and $\mathcal{M}_{\widehat{\mathbf{W}}^m}$ be the global sub-model and the local sub-model for the client $m$, respectively. It can be formulated as
\begin{equation}\label{myeq5}
\mathcal{M}_{\mathbf{W}^m}=\mathcal{F}_m\left(\mathcal{M}_{\widetilde{\mathbf{W}}^m}, \mathcal{M}_{\widehat{\mathbf{W}}^m}\right)
\end{equation}
where $\mathcal{F}_m\left(\cdot\right)$ is an aggregate function. Replace Eq. (\ref{myeq2}) with Eq. (\ref{myeq5}) to get the following rewritten global objective as

\begin{equation} \label{myeq6}
\arg \min _{\mathbf{W}^*, \mathit{~W}} \sum_{\mathrm{t}=1}^{\mathrm{m}} \mathcal{G}_{\mathbf{W}^*}\left(\sum_{\mathrm{i}=1}^{\mathrm{n}_{\mathrm{t}}} \mathcal{L}_{\mathrm{t}}\left(\mathcal{F}_{\mathrm{t}}\left(\mathcal{M}_{\widetilde{\mathbf{W}}^t}\left(\mathrm{x}_{\mathrm{i}}\right), \mathcal{M}_{\widehat{\mathbf{W}}^t}\left(\mathrm{x}_{\mathrm{i}}\right)\right)\right)\right)
\end{equation}
where $\mathit{W}=\{\{{\widetilde{\mathbf{W}}^1},{\widehat{\mathbf{W}}^1}\},\{{\widetilde{\mathbf{W}}^2},{\widehat{\mathbf{W}}^2}\},\cdots,\{{\widetilde{\mathbf{W}}^m},{\widehat{\mathbf{W}}^m}\}\}$. To widen the distance between two submodels of clients, a regularization term is introduced into Eq. (\ref{myeq6}) as
\begin{equation}
\mathcal{E}\left(\mathbf{W}^t\right)=\operatorname{dist}\left(\mathcal{M}_{\widetilde{\mathbf{W}}^t}, \mathcal{M}_{\widehat{\mathbf{W}}^t}\right)=\exp \left(-\frac{\left\|\widetilde{\mathbf{W}}^t-\widehat{\mathrm{\mathbf{W}}}^t\right\|_{\mathrm{F}}}{\left\|\widetilde{\mathbf{W}}^t\right\|_{\mathrm{F}}+\left\|\widehat{\mathbf{W}}^t\right\|_{\mathrm{F}}}\right)
\end{equation}
where $\operatorname{dist}\left(
\mathit{A},\mathit{B}\right)$ measures the difference between A and B; $\left\|\mathbf{A}\right\|_{\mathrm{F}}=\sqrt{\sum_{i=1}^{m} \sum_{j=1}^{n}\left|a_{i j}\right|^{2}}$ calculates the Fibonacci norm of the matrix $\mathbf{A}$ which aims to collect the difference between all parameters between two identical models, correspondingly; $\exp\left(\cdot\right)$ is the exponential function.

Thus, the objective function is implemented by combining Eqs. (6) and (7), where $\alpha\geq0$ is a balance parameter.
\begin{equation}\label{myeq8}
\begin{split}
\arg \min _{\mathbf{W}^*, \mathit{~W}} \sum_{\mathrm{t}=1}^{\mathrm{m}} \mathcal{G}_{\mathbf{W}^*} \left( \sum_{\mathrm{i}=1}^{\mathrm{n}_{\mathrm{t}}} \mathcal{L}_{\mathrm{t}}\left(\mathcal{F}_{\mathrm{t}}\left(\mathcal{M}_{\widetilde{\mathbf{W}}^t}\left(\mathrm{x}_{\mathrm{i}}\right), \mathcal{M}_{\widehat{\mathbf{W}}^t}\left(\mathrm{x}_{\mathrm{i}}\right)\right)\right)\right) \\
+\alpha \operatorname{dist}\left(\mathcal{M}_{\widetilde{\mathbf{W}}^t}, \mathcal{M}_{\widehat{\mathbf{W}}^{\mathrm{t}}}\right)
\end{split}
\end{equation}

\noindent\textit{The proposed FecMap combines the strategy of MPP}. Denote by $\mathbf{F}^*$ and $\mathbf{F}^{'}$ the shareable features and the privacy features, respectively. By MPP, it can be cast as
\begin{equation} \label{myeq9}
    \mathbf{F}^*=\mathbf{g}_i, \mathbf{F}^{'}=\mathbf{s_\mathit{i}}\oplus \mathbf{c_\mathit{i}}
\end{equation}
where student grades $\mathbf{g}_i=[g_1,g_2,\cdots,g_k]$ are employed as shareable features, while student personal information $\mathbf{s_\mathit{i}}=[s_1,s_2,\cdots,s_m]$ and course description $\mathbf{c_\mathit{i}}=[c_1,c_2,\cdots,c_n]$ are used as not-allowed features in this study. 

Finally, substitute Eq. (\ref{myeq9}) into Eq. (\ref{myeq8}) and then obtain the following global objective problem.
\begin{equation}
\begin{split}
\arg \min _{\mathbf{W}^*, \mathrm{~W}} \sum_{\mathrm{t}=1}^{\mathrm{m}} \mathcal{G}_{\mathbf{W}^*}\left(\sum_{\mathrm{i}=1}^{\mathrm{n}_{\mathrm{t}}} \mathcal{L}_{\mathrm{t}}\left(\mathcal{F}_{\mathrm{t}}\left(\mathcal{M}_{\widetilde{\mathbf{W}}^t}\left(\mathbf{F}_i^*\right), \mathcal{M}_{\widehat{\mathbf{W}}^t}\left(\mathbf{F}_i^*\right), \mathcal{H}^{t}\left(\mathbf{F}_i^{'}\right)\right)\right)\right)\\
+\alpha \operatorname{dist}\left(\mathcal{M}_{\widetilde{\mathbf{W}}^t}, \mathcal{M}_{\widehat{\mathbf{W}}^{\mathrm{t}}}\right)
\end{split}
\end{equation}
where $\mathcal{M}_{\mathbf{\widetilde{W}}^t}$, $\mathcal{M}_{\mathbf{\widehat{W}}^t}$, and $\mathcal{H}^{t}$ are fulfilled by the shared sub-model, the not-shared sub-model, and the local head.
\\\\
% \subsection{The FecMap Algorithm}
3.3 \quad The FecMap Algorithm

\begin{figure}
\centering
\includegraphics[width=0.5\textwidth]{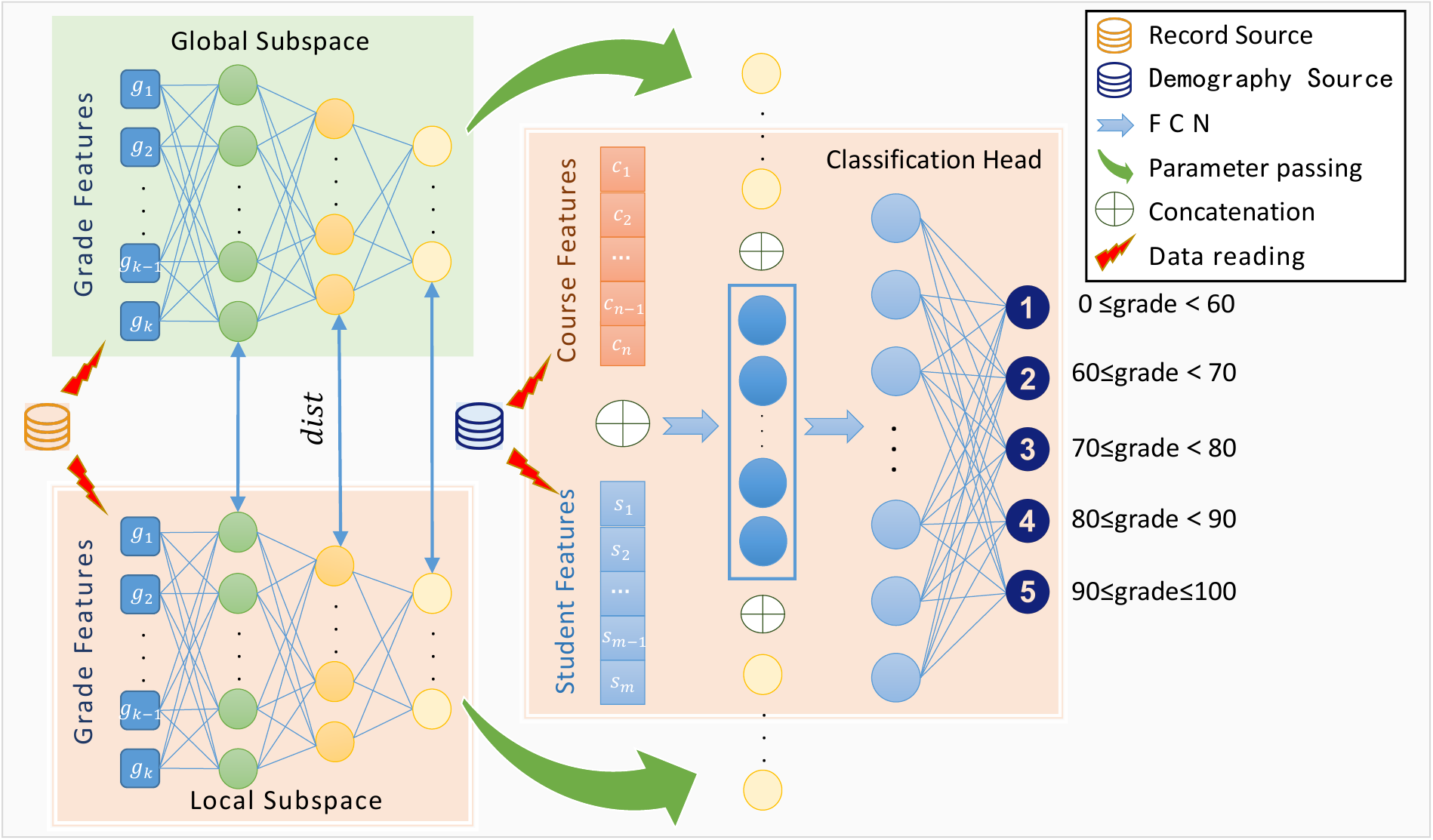}
\caption{Schematic diagram of the FecMap client neural network model. Circle points are network nodes; green box is sharable; red boxes are non-sharable. $\mathbf{g}_i$ is the grade, and $dist$ is the distance between the two networks. The learned representations are composed of student features $\mathbf{s_\mathit{i}}$ and course features $\mathbf{c_\mathit{i}}$. Outputs are the levels of the course scores.}
\label{Network}
\end{figure}

\noindent The proposed FecMap algorithm is implemented in the commonly used iteration manner, as shown in Algorithm 1.

\textbf{Client Update.} The task for each client is to train the LOP classifier on their respective datasets. In the $k$-th round of FL communication, the network weights are updated by
\begin{equation}
    \left(\widetilde{\mathbf{W}}^{t}_{(\mathrm{k})}, \widehat{\mathbf{W}}^{t}_{(\mathrm{k})}\right) \leftarrow \operatorname{SGD}\left(\mathbf{W}^*_{(\mathrm{k}-1)}, \widehat{\mathrm{\mathbf{W}}}^{t}_{(\mathrm{k}-1)}, \mathcal{H}^{t}_{(\mathrm{k-1})}\right)
\end{equation}
where $\operatorname{SGD}\left(\cdot\right)$ is the stochastic gradient descent and $\mathrm{W}^*_{(\mathrm{k-1})}$ is the global model from the previous round.
 
\textbf{Server Update.} After client updating, all clients upload the global part $\widetilde{\mathrm{W}^{t}}_{(\mathrm{k})}$ to the server. The server then performs an average of all clients and returns the new global model to clients. The server model is then computed by
\begin{equation}
    \mathbf{W}^*_{(k+1)}=\frac{1}{m} \sum_{t=1}^{m} \widetilde{\mathbf{W}}^{t}_{(\mathrm{k})}
\end{equation}

\section{Model Implementation and Evaluation}

% \subsection{An Online FecMap Implementation}
4.1 \quad An Online FecMap Implementation

\begin{figure}[t]
\centering
\includegraphics[width=0.44\textwidth]{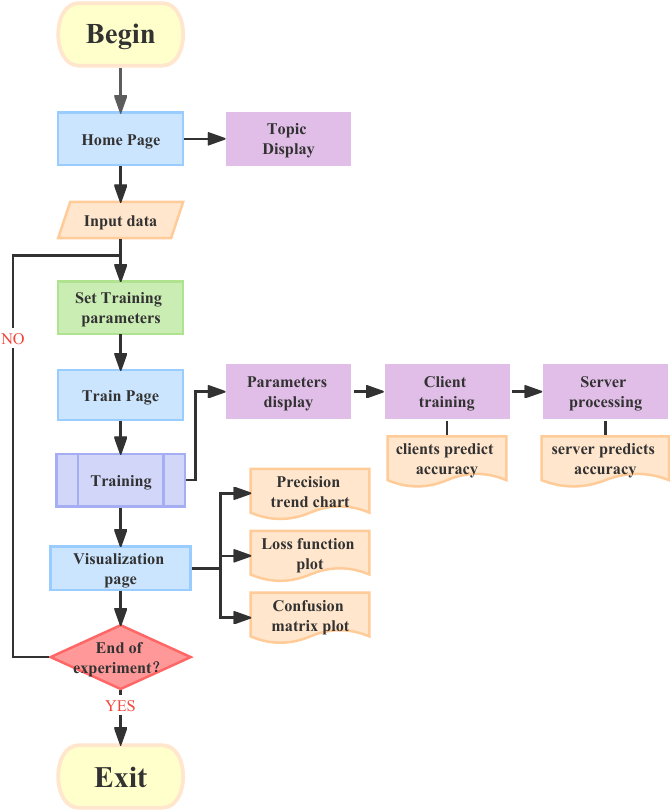}
\caption{The Online FecMap flowchart.}
\label{Web_framework}
\end{figure}

\noindent To use the proposed FecMap for LOP in multiple institutions, we have developed an online FL platform by employing the open-source technology of Streamlit. The online workflow is shown in Fig. \ref{Web_framework}, including the interfaces of data loading, parameter setting, model training, results visualization, and result evaluation. Our experiments are conducted on multiple computers and a server via this online version.
\begin{algorithm}[t]
	\renewcommand{\algorithmicrequire}{\textbf{Require:}}
	\caption{FecMap Algorithm}
	\label{alg:1}
	\begin{algorithmic}[1]
		\REQUIRE
		\quad Participation rate $\gamma$; step size $\alpha$; the number of local updates $\tau$; the number of communication rounds $T$\\
		\textbf{Initialization} \quad $\mathbf{W}^*_0$; $\mathbf{\mathcal{H}}_0^1, \cdots, \mathbf{\mathcal{H}}_0^n$; $\mathbf{\widehat{W}}^0_0, \cdots, \mathbf{\widehat{W}}^n_0$
		\FOR{$t=1,2, \ldots, T$}
		\STATE
        Server receives a batch of clients $L^t$ of size $\gamma n$
        \\Server sends current global model $\mathbf{W}^*_{t-1}$ to clients
        \FOR{each client $i$ in $L^t$}
        \STATE
        Client $i$ initializes 
        \\$\widetilde{\mathbf{W}}_{t}^{\mathrm{i}} \leftarrow \mathbf{W}^*_{t-1}, \widehat{\mathbf{W}}_{t}^{\mathrm{i}} \leftarrow \widehat{\mathbf{W}}_{t-1}^{i}, \mathcal{H}_{t}^{i} \leftarrow \mathcal{H}_{t-1}^{i}$
        \\Updates the local sub-model $\widehat{\mathbf{W}}_{t}^{i}$ and local head $\mathcal{H}_{t}^{i}$
        \FOR{$s=1$ to $\tau$}
        \STATE
        $\left(\widetilde{\mathbf{W}}^{i}_{t+1,s}, \widehat{\mathbf{W}}^{i}_{t+1,s}\right) \leftarrow \operatorname{SGD}\left(\mathbf{W}^*_{t,s}, \widehat{\mathrm{\mathbf{W}}}^{i}_{t,s}, \mathcal{H}^{i}_{t,s}, \alpha\right)$\\
        \ENDFOR
        \\Client $i$ sends updated global model $\widetilde{\mathbf{W}}_{t,\tau}^{i}$ to server
        \ENDFOR
        \FOR{each client $i$ not in $L^t$}
        \STATE
        Set $\mathbf{\widehat{W}}^i_{t,\tau}\leftarrow\mathbf{\widehat{W}}^i_{t-1,\tau}$, ${\mathcal{H}}_{t, \tau}^{i} \leftarrow {\mathcal{H}}_{t-1, \tau}^{i}$
        \ENDFOR
        \\Server computes the new global model as:  \\$\mathbf{W}^*_{t+1}=\frac{1}{m} \sum_{i=1}^{m} \widetilde{\mathbf{W}}_{t}^{i}$
        \ENDFOR
	\end{algorithmic}  
\end{algorithm}
\\\\
% \subsection{The Used Dataset}
4.2 \quad The Used Dataset

\noindent To evaluate the FecMap model, we collected our university's undergraduate student performance records in the years $2015 - 2020$. Each student has two semesters a year, each free for course selection. A final examination is usually administered to assess the student's learning outcomes at the end of a course. Most results are given in $\left(0 - 100\right)$ or grades $\left(A - E\right)$, and other forms are removed, such as "pass" and "fail".

In this paper, records from three majors, i.e., Computer Science and Technology (CST), Software Engineering (SE), and Electronic Information Engineering (EIE), are selected for the experiment. Detailed information, including sample sizes and basic classification statistics for each dataset, is shown in Table \ref{dataset}. In addition, student characteristics include age, class, gender, country, and ethnicity. Course characteristics include total course hours and weekly course hours, credits, category, textbook used, course examination method, and course offering unit. All attributes are converted to numerical values and then normalized between 0 and 1.
% dataset
\doublerulesep 0.1pt
\begin{table*}[t]
\begin{footnotesize}
\caption{Data set information, where the values in brackets are the ratios.} 
\label{dataset}
\begin{tabular}{p{1.5cm}p{1.5cm}p{1.5cm}p{1.5cm}p{1.5cm}p{1.5cm}p{1.5cm}p{1.5cm}p{1.5cm}}
\hline\hline\noalign{\smallskip}
     \textbf{Major} & \textbf{Student} & \textbf{Course} & \textbf{Record} & \textbf{0$\le$grade$<$60} & \textbf{60$\le$grade$<$70} & \textbf{70$\le$grade$<$80}& \textbf{80$\le$grade$<$90}& \textbf{90$\le$grade$<$100} \\
\noalign{\smallskip} \hline
        CST & 1463 & 44 & 46512 & 1874(4.0\%) & 7434(15.9\%) & 10755(23.1\%) & 15720(33.7\%)& 10729(23.0\%) \\ 
        SE & 1621 & 46 & 37952 & 1445(3.8\%) & 7050(18.5\%) & 9437(24.8\%) & 13048(34.3\%)& 6972(18.3\%) \\
        EIE & 1937 & 43 & 49546 & 1917(3.8\%) & 9128(18.4\%) & 10655(21.5\%) & 16715(33.7\%)& 11131(22.4\%) \\
\hline\hline
\end{tabular}
\end{footnotesize}
\end{table*}
\\\\
% \subsection{Experimental Settings}
4.3 \quad Experimental Settings

\noindent We conducted the experiments via the following settings.

\textbf{Data partitioning.} The dataset in hand is equally divided for all clients with the same class distribution. The dataset allocated to each client is then partitioned into a training set and a test set by a ratio of 7:3.

\textbf{Model parameters.} FecMap in this experiment employs MLP for $\mathcal{H}$, ReLU for the activation function, the softmax function for classification, and the cross-entropy for loss computation. There are in total 8 fully connected layers in the overall network. Concretely, the global part has $dim_{in}$-22-11-6 units; the local part has $dim_{in}$-22-11-6 units; the classification head has 27-20-13-5 units. The feature dimension determines the model input $dim_{in}$.

\textbf{Model training.} FecMap randomly initializes all models for $T=500$ communication rounds of training. Clients are randomly selected at a rate of $\gamma=0.1$ to participate in federated machine learning. The local model in clients performs 15 SGD in each local update, where the learning rate and the moment are set to 0.01 and 0.5, respectively.

\textbf{Model comparisons.} Five state-of-the-art FL methods, FedAvg \cite{mcmahan2017communication}, FedProx \cite{li2020federated2}, LG-Fed \cite{liang2020think}, FedPer \cite{arivazhagan2019federated} and FedRep \cite{collins2021exploiting} are meanwhile evaluated for comparison.
\\\\
4.4 \quad Experiment Results
% \subsection{Experiment Results}
\begin{figure*}
\centering
\subfigure{
    \begin{minipage}[b]{0.23\linewidth} %0.23为minipage的宽度，可以调节子图间的距离
    \includegraphics[width=4cm]{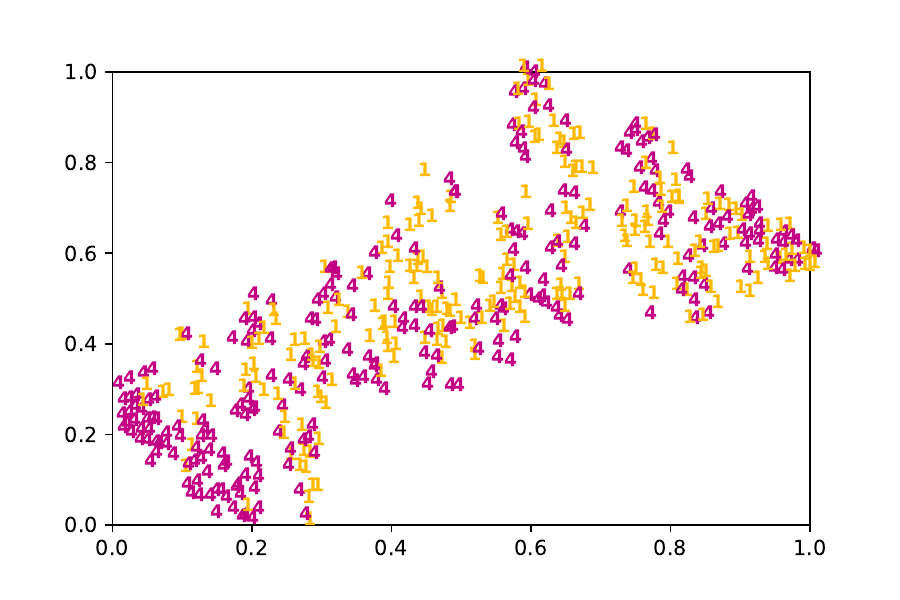}\vspace{1pt} %图片的宽度、路径和垂直间距
    \includegraphics[width=4cm]{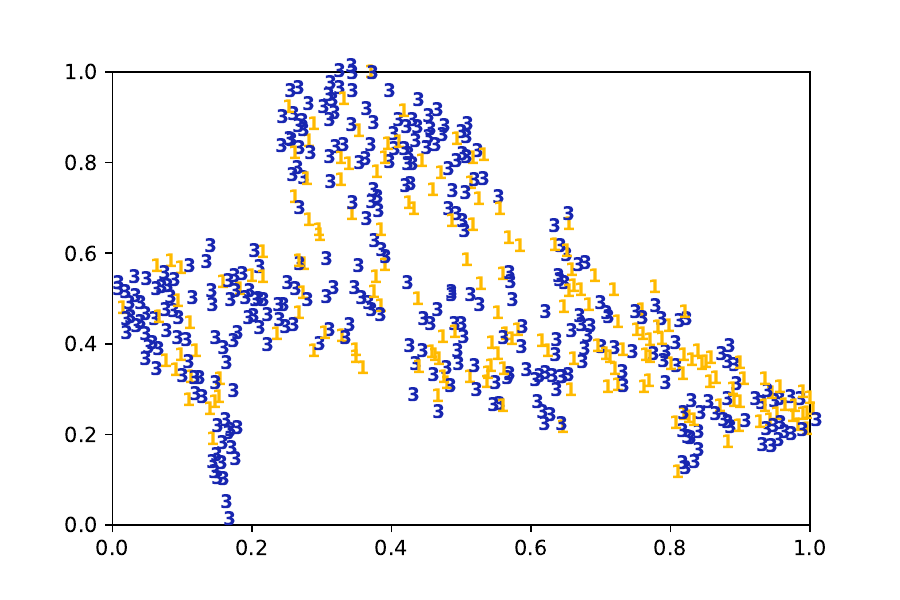}\vspace{1pt}
    %\vspace要紧跟在对应的includegraphics，不然得不到想要的结果
    \includegraphics[width=4cm]{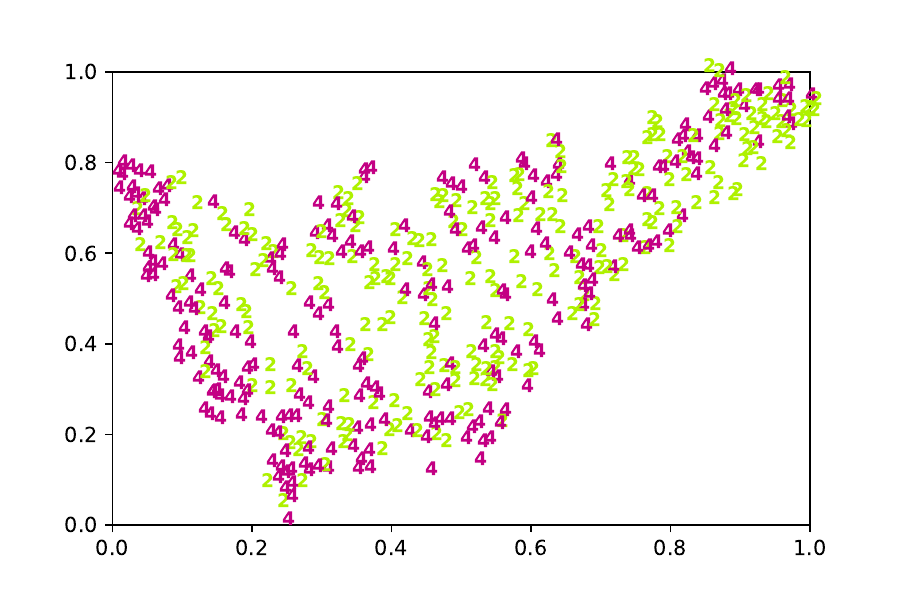}\vspace{1pt}
    \end{minipage}
}
\subfigure{
    \begin{minipage}[b]{0.23\linewidth}
    \includegraphics[width=4cm]{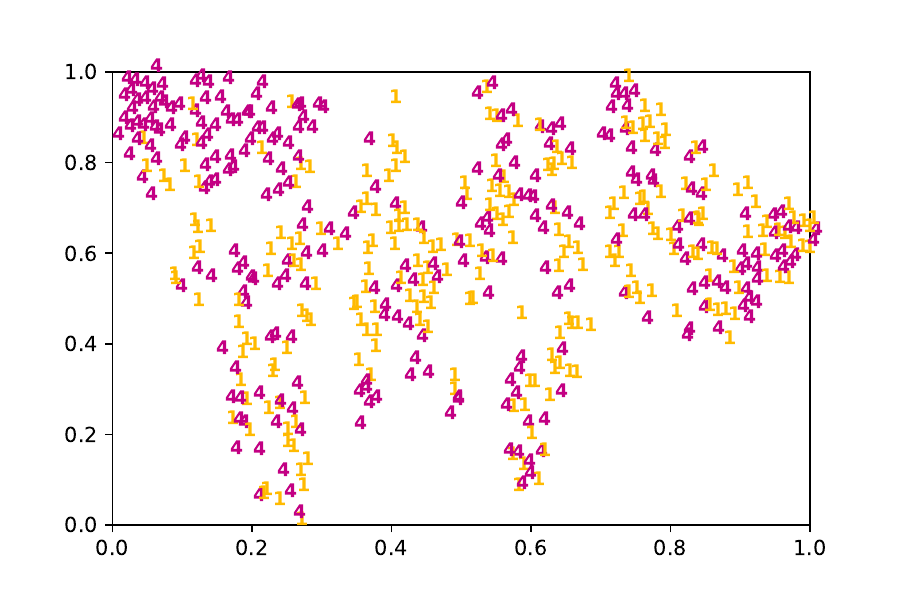}\vspace{1pt} 
    \includegraphics[width=4cm]{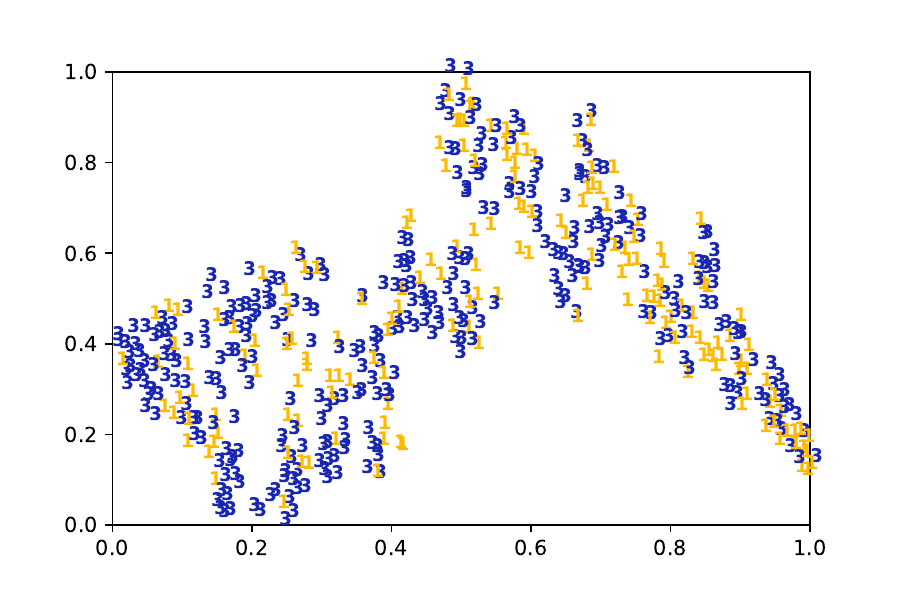}\vspace{1pt}
    \includegraphics[width=4cm]{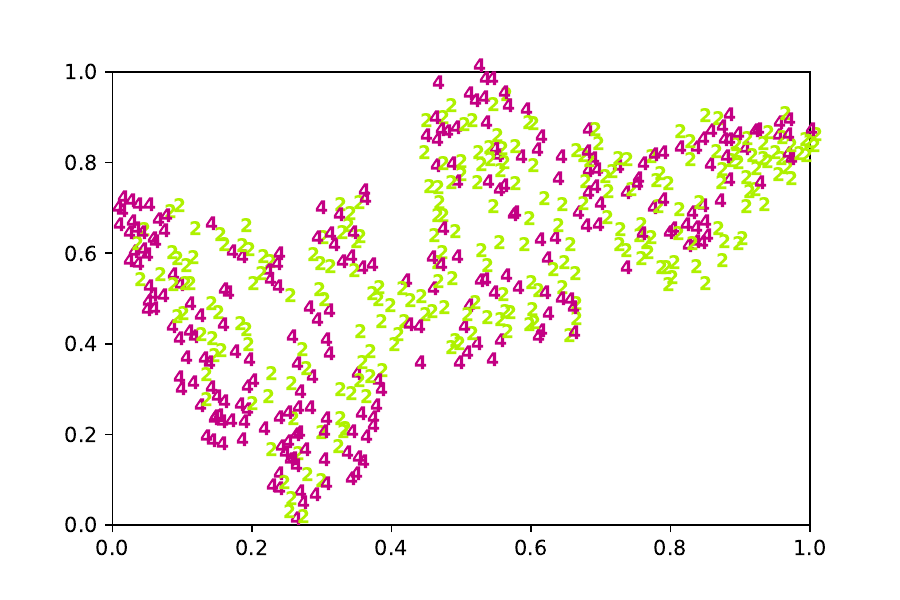}\vspace{1pt}
    \end{minipage}
}
\subfigure{
    \begin{minipage}[b]{0.23\linewidth}
    \includegraphics[width=4cm]{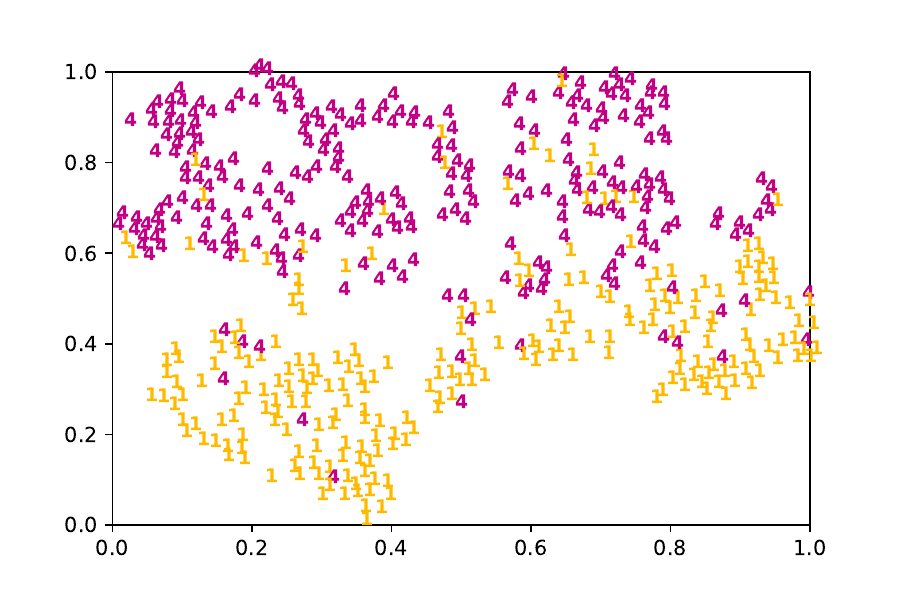}\vspace{1pt} 
    \includegraphics[width=4cm]{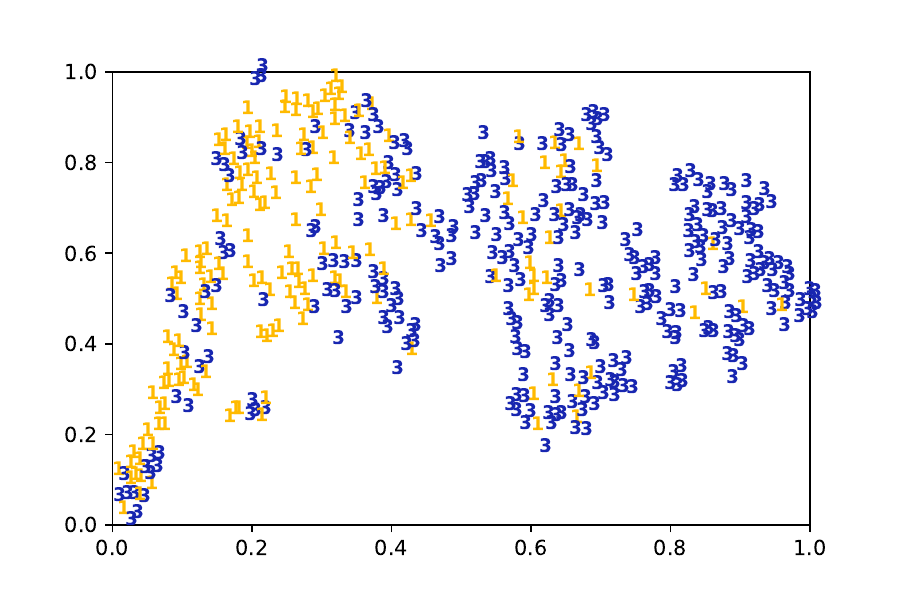}\vspace{1pt}
    \includegraphics[width=4cm]{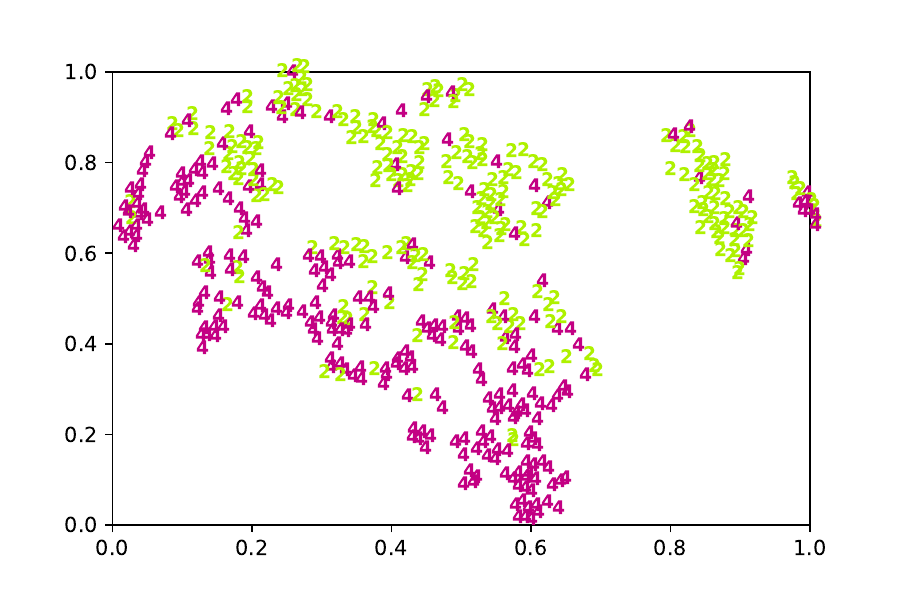}\vspace{1pt}
    \end{minipage}
}
\subfigure{
    \begin{minipage}[b]{0.23\linewidth}
    \includegraphics[width=4cm]{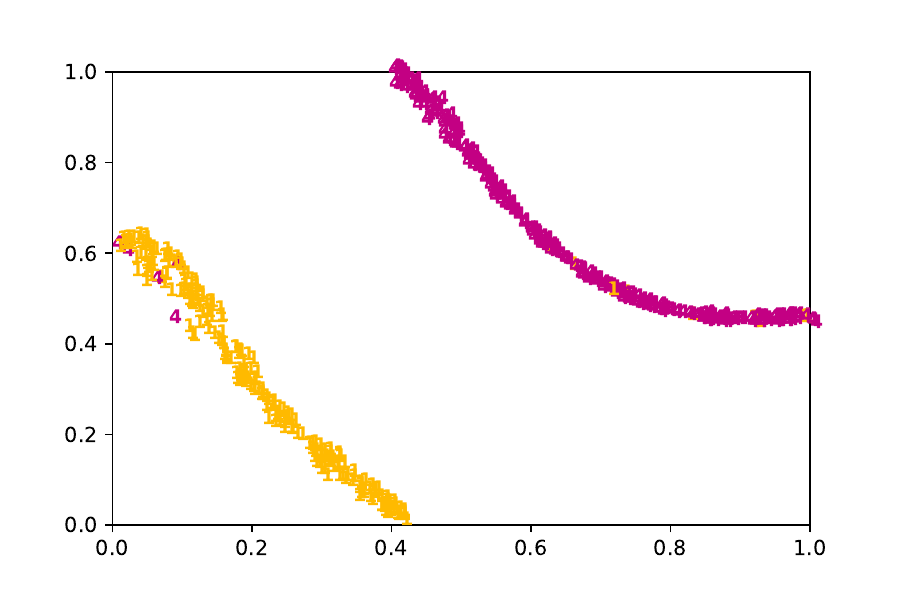}\vspace{1pt} 
    \includegraphics[width=4cm]{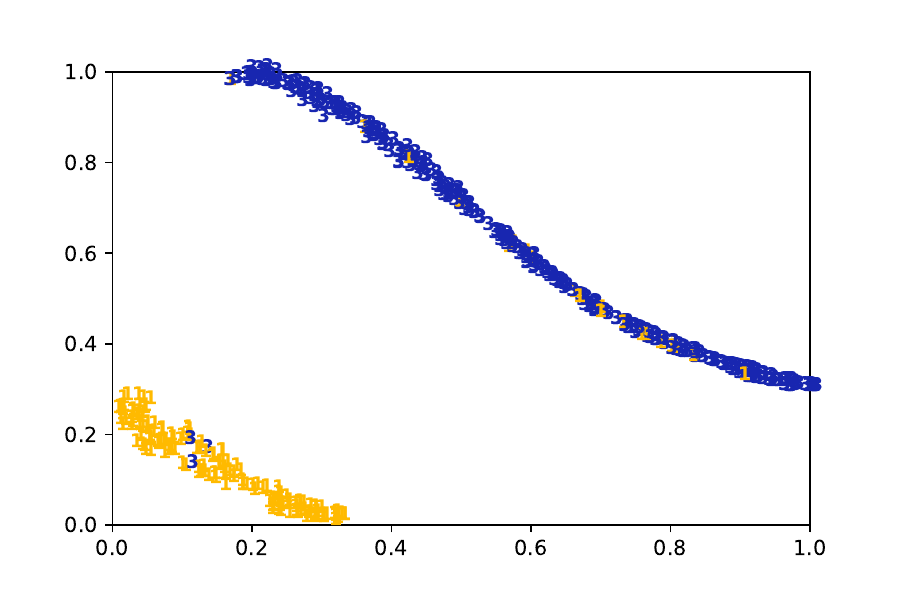}\vspace{1pt}
    \includegraphics[width=4cm]{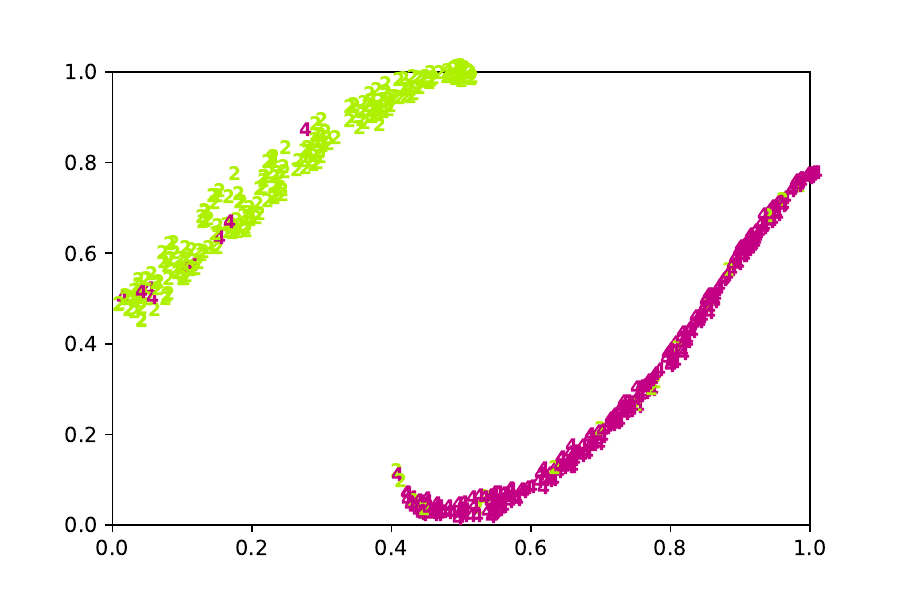}\vspace{1pt}
    \end{minipage}
}
\caption{Visualization of data representation. The first column is the global representations, the second is the local representations, the third is the combined representations, and the fourth is the discriminative representations. Each row represents a client, and each color represents a category.}
\label{data visualisation}
\end{figure*}
\\\\
% \subsubsection{Data Representation Visualization}
4.4.1 \quad Data Representation Visualization

\noindent In order to observe the discriminative power of the model, we visualized the representations by the last layer of the local and global parts in Fig. \ref{Network} for two classes. The resulting 8D subspace features are projected into 2D representations by using TSNE \cite{van2008visualizing} for visualization. Fig. \ref{data visualisation} shows the global representation, the local representation, and the combined representation from the 1st to the 3rd column, respectively. Besides, the discriminative representation from the layer before the softmax layer is also displayed in the 4th column.

As shown in Fig. \ref{data visualisation}, the results of three clients are shown, where different colors represent the categories labeled with numbers. From the results, the global representations are much mixed, while the local representations potentially have improvements. Finally, the combination of both local and global representations is more discriminative for distinguishing the two classes clearly.
\\\\
% \subsubsection{Prediction Results}
4.4.2 \quad Prediction Results

\noindent In order to test the prediction accuracy of the model for student performance, the proposed FecMap and these comparison methods are tested on each of the three datasets. The client model is evaluated on the local test set of all clients, and then the results are averaged to obtain the final accuracy. Table \ref{Accuracy comparison} lists all results.

\doublerulesep 0.1pt
\begin{table}[t] 
\begin{footnotesize}
\caption{Accuracy comparison of various methods.}
\label{Accuracy comparison}
\begin{tabular}{p{2cm}p{1.6cm}p{1.6cm}p{1.5cm}}
\hline\hline\noalign{\smallskip}
    \textbf{Methods} & \textbf{CST} & \textbf{SE}& \textbf{EIE}\\
\noalign{\smallskip} \hline
            FedAvg (\cite{mcmahan2017communication}) & 80.46 & 76.82 & 77.86 \\
            FedProx (\cite{li2020federated2}) & 79.23 & 78.23 & 78.19 \\
            LG-Fed (\cite{liang2020think}) & 76.50  & 75.85 & 78.51 \\
            FedPer (\cite{arivazhagan2019federated}) & 81.58 & 78.29 & 77.52 \\
            FedRep (\cite{collins2021exploiting}) & 81.87 & 83.95 & 82.99 \\
\noalign{\smallskip} \hline
            \textbf{FecMap(Ours)} & \textbf{83.10} & \textbf{84.92} & \textbf{85.41} \\
\hline\hline
\end{tabular}
\end{footnotesize}
\end{table}

As shown in Table \ref{Accuracy comparison}, FecMap obtains higher prediction performance than the state-of-the-art FL methods. Specifically, on the CST dataset, the accuracy outperformed the FedAvg, FedProx, LG-Fed, FedPer and FedRep methods by 2.64\%, 3.87\%, 6.6\%, 1.52\% and 1.23\%, respectively. Similar results can be observed on the SE and EIE datasets. It is demonstrated that the proposed model outperforms existing FL methods overall and performs well in the task of LOP. The results imply that FecMap can benefit from the proposed LSL and MPP.

\begin{figure*}[t]
\begin{minipage}{0.49\linewidth}	% linewidth就是栏宽
    \subfigure{
        \includegraphics[width=0.475\textwidth,height=1.12in]{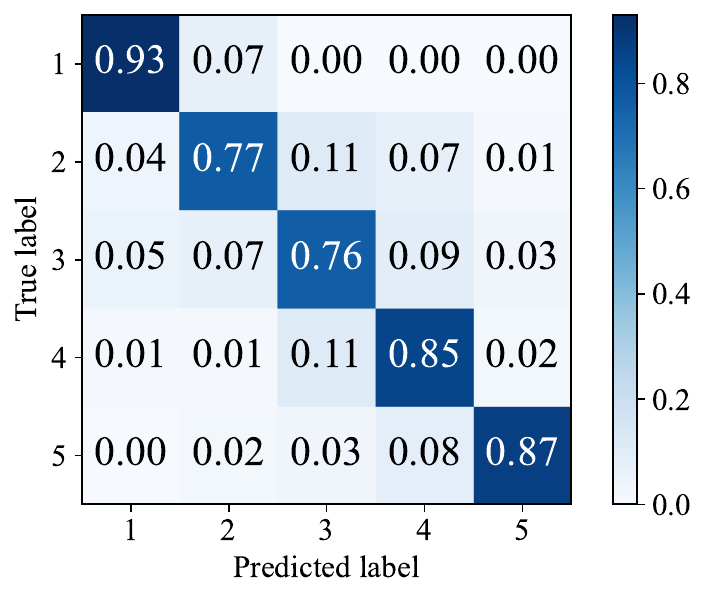}
    }
    \subfigure{
        \includegraphics[width=0.475\textwidth,height=1.12in]{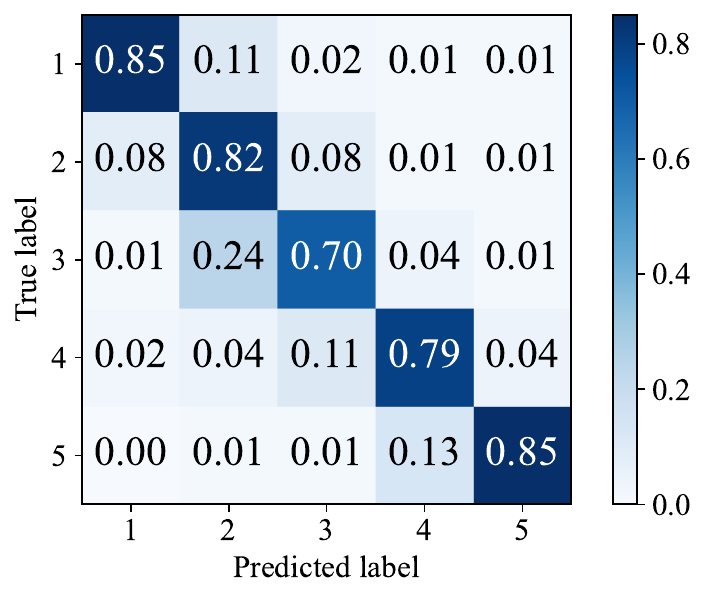}
    }
    \caption{Confusion matrix for FecMap (left) and FedRep (right).}
    \label{Confusion matrix}
\end{minipage}
\hfill  
\begin{minipage}{0.49\linewidth}	% linewidth就是栏宽
    \subfigure{
        \includegraphics[width=0.475\textwidth,height=1.03in]{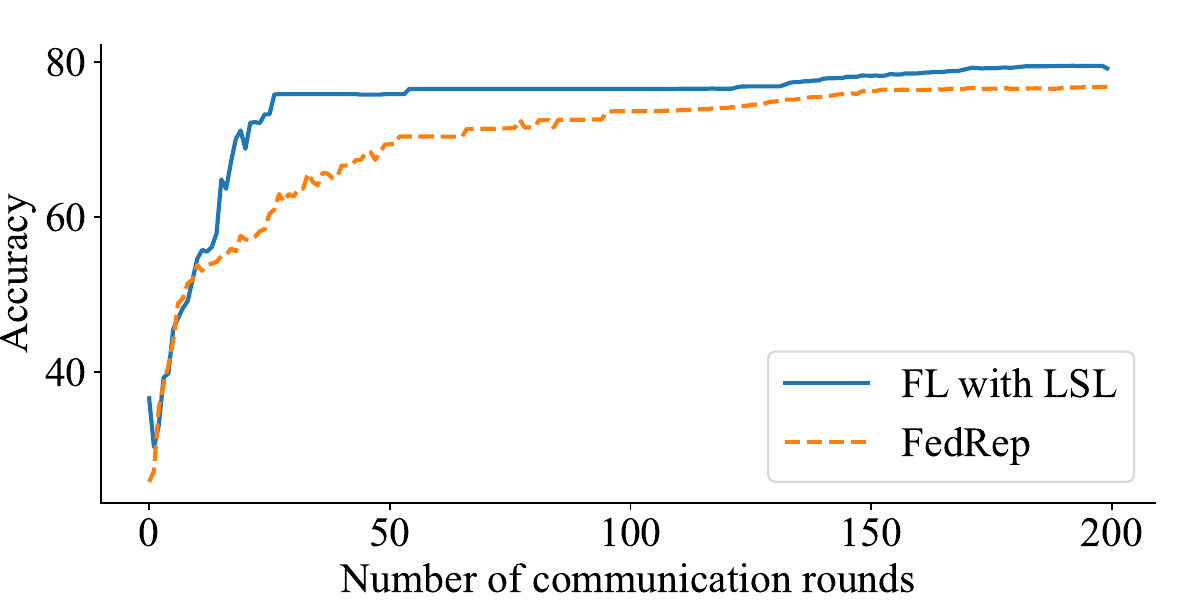}
    }
    \subfigure{
        \includegraphics[width=0.475\textwidth,height=1.03in]{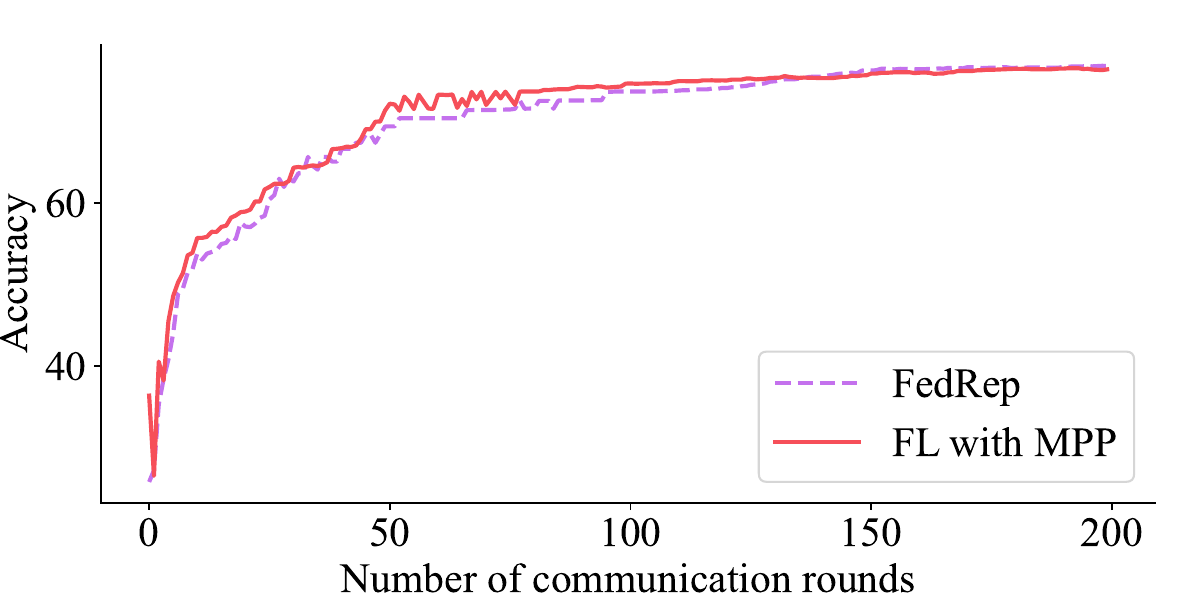}
    }
    \caption{The accuracy of FecMap-LSL and FecMap-MPP against the number of communications, compared to FedRep.}
    \label{myabl}
\end{minipage}
\end{figure*}

Fig. \ref{Confusion matrix} compares the confusion matrixes achieved from FecMap and FedRep on the test sets. In comparison to the FedRep framework, the proposed method achieves higher prediction accuracy in all five categories. In particular, our FecMap achieves an accuracy of 93\% in detecting students with course grades below 60 (grade 1), while FedRep was only 85\%. This result shows that this method is practical and feasible in an academic early warning system. For the prediction of grades 3, 4 and 5, FecMap also delivers higher accuracy than FedRep, although there are more errors in grade 2. In addition, FecMap achieves excellent performance in distinguishing between grade 1 and grade 5.
% % figure
% \begin{figure*}[t]
% \begin{center}{
% \includegraphics[width=0.35\textwidth]{figure/Confusion matrix_FecMap.pdf}
% }
% \hspace{.3in}
% {
% \includegraphics[width=0.35\textwidth]{figure/Confusion matrix_FedRep.pdf}
% }
% \caption{Confusion matrix for FecMap (left) and FedRep (right).}
% \label{Confusion matrix}
% \end{center}
% \end{figure*}
% % figure
% \begin{figure*}[t]
% \begin{center}{
% \includegraphics[width=0.35\textwidth]{figure/Round convergence_LSL.pdf}
% }
% \hspace{.3in}
% {
% \includegraphics[width=0.35\textwidth]{figure/Round convergence_MPP.pdf}
% }
% \caption{The accuracy of FecMap-LSL and FecMap-MPP against the number of communications, compared to FedRep.}
% \label{myabl}
% \end{center}
% \end{figure*}
% figure
\begin{figure*}[t]
\begin{center}{
\includegraphics[width=0.4\textwidth]{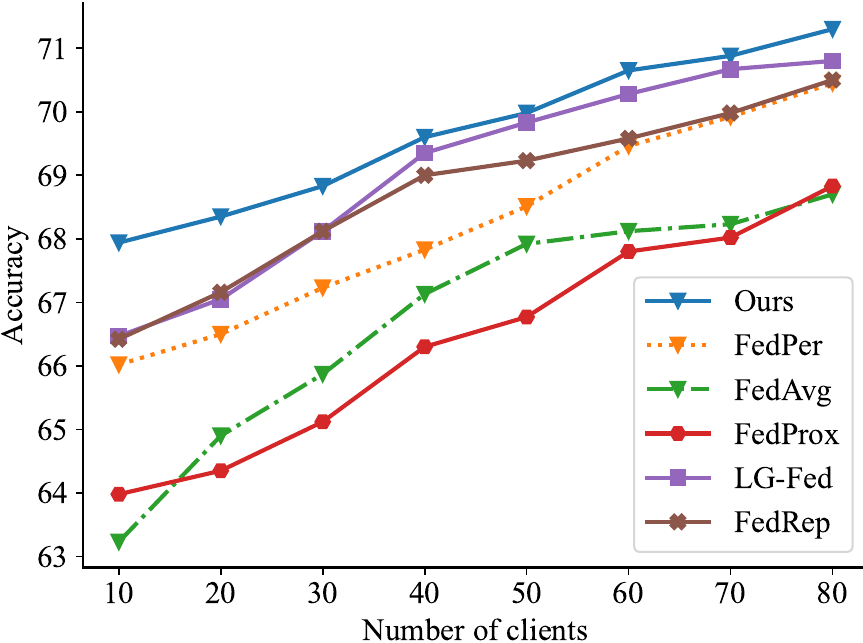}
}
\hspace{.4in}
\includegraphics[width=0.4\textwidth]{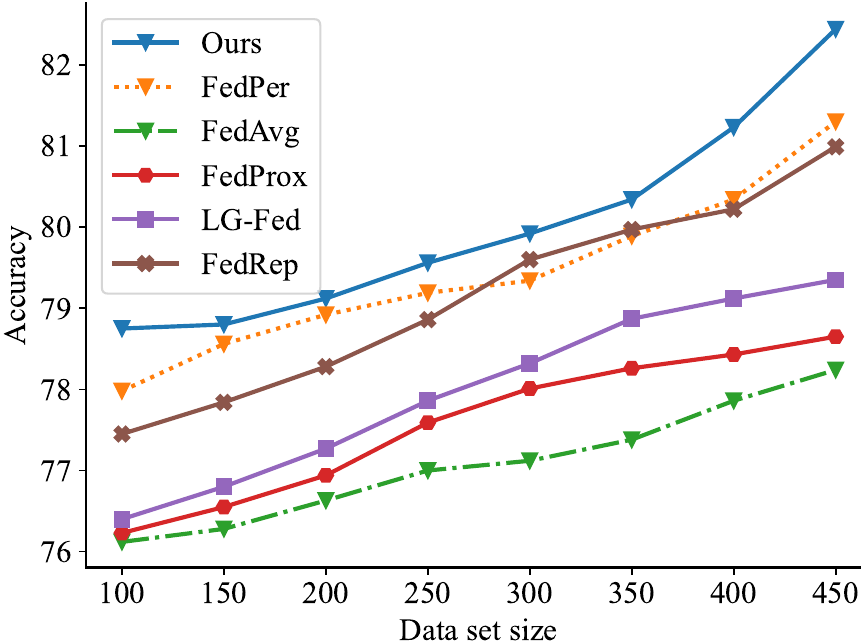}
\caption{Parameter discussion on the number of clients $n$ (left) and the data set size $d$ (the number of samples per client, right) in the proposed model and other comparison methods.}
\label{Prediction results}
\end{center}
\end{figure*}
% loss
% \begin{figure}[t]
% \begin{center}{
% \includegraphics[width=0.3\textwidth]{figure/loss_curve.pdf}
% }
% {
% \includegraphics[width=0.3\textwidth]{figure/accuracy_curve.pdf}
% }
% \caption{Loss function (top) and accuracy (bottom) for FecMap in the case study with 20 communication rounds.}
% \label{Loss function}
% \end{center}
% \end{figure}
\begin{figure}[t]
\begin{minipage}{1\linewidth}	% linewidth就是栏宽
    \subfigure{
        \includegraphics[width=0.47\linewidth,height=1.14in]{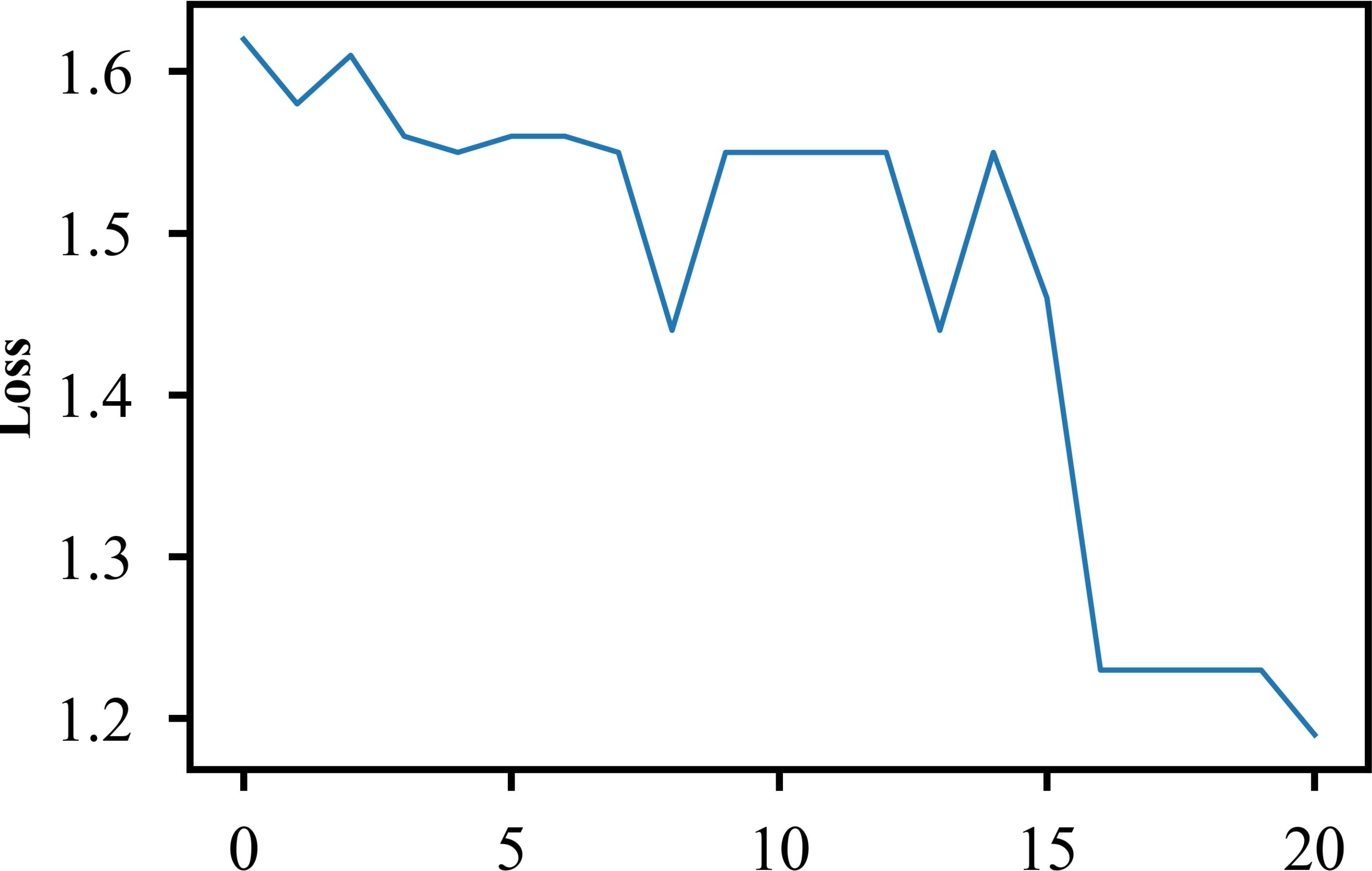}
    }
    \subfigure{
        \includegraphics[width=0.47\linewidth,height=1.14in]{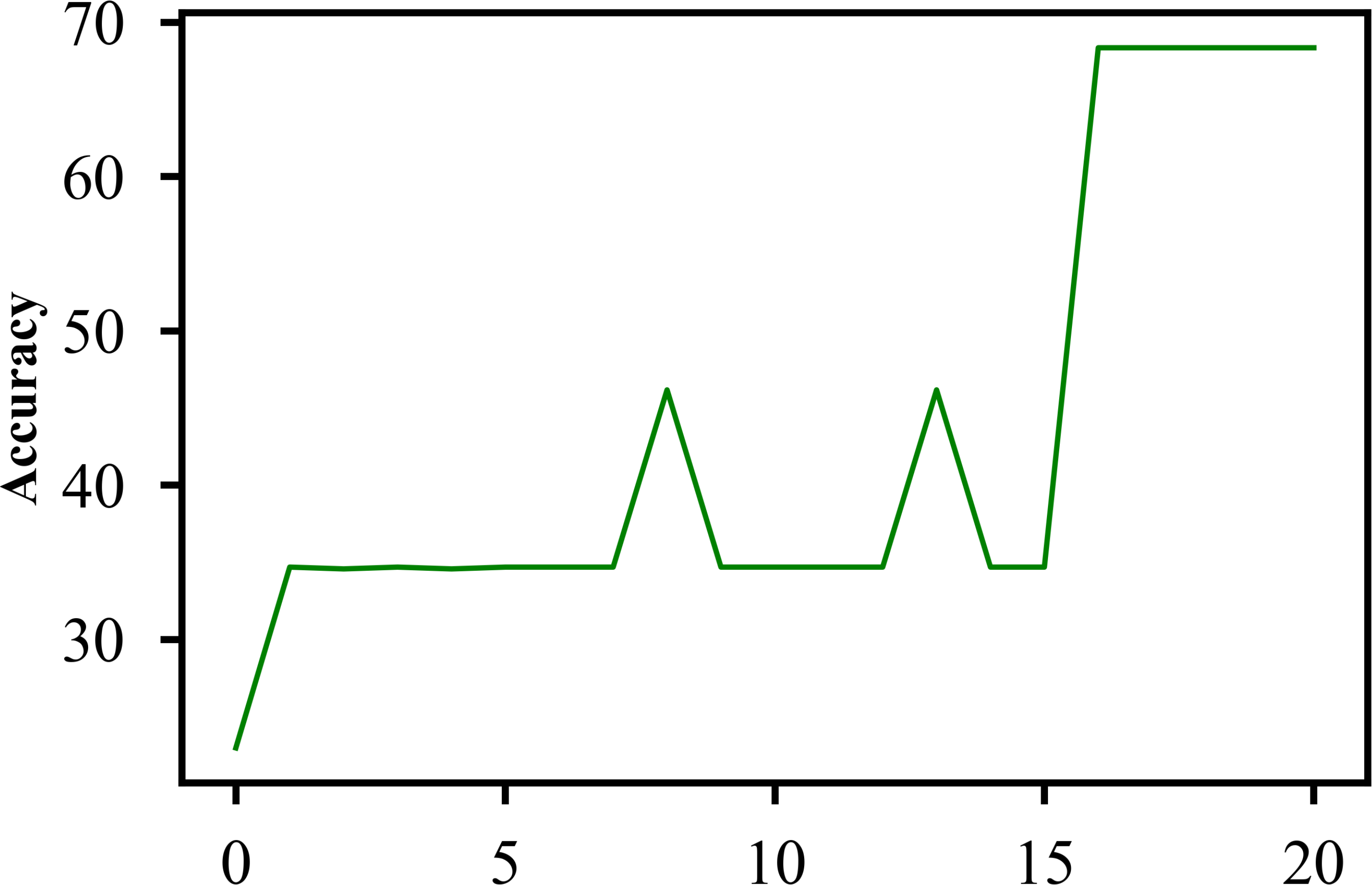}
    }
    \caption{Loss function (left) and accuracy (right) for FecMap in the case study with 20 communication rounds.}
    \label{Loss function}
\end{minipage}
\end{figure}
\\\\
% \subsubsection{Ablation Study}
4.4.3 \quad Ablation Study

\noindent To evaluate the impacts of the introduced LSL and MPP, we evaluated the FecMap model with only MPP (FecMap-MPP) and the FecMap model with only LSL (FecMap-LSL) on the three datasets, respectively. Besides, we compared with FedRep as it is our base model. Fig. \ref{myabl} shows the prediction performance of FecMap-LSL vs. FedRep and FecMap-MPP vs. FedRep. The results show that FecMap can benefit from the introduced LSL and MPP. The final accuracy is listed in Table \ref{Ablation study}, showing that both LSL and MPP can improve the performance of FedRep, where LSL seems more powerful than MPP. Overall, FecMap benefits from MPP and LSL to achieve the best results.
\doublerulesep 0.1pt
\begin{table}[t]
\begin{footnotesize}
\caption{Ablation study} 
\label{Ablation study}
\begin{tabular}{p{2cm}p{1.6cm}p{1.6cm}p{1.5cm}}
\hline\hline\noalign{\smallskip}
    \textbf{Methods} & \textbf{CST} & \textbf{SE}& \textbf{EIE}\\
\noalign{\smallskip} \hline
        FecMap - LSL & 82.58 & 84.56 & 85.30 \\
        FecMap - MPP & 82.01 & 84.36 & 83.18 \\
        FedRep & 81.87 & 83.95 & 82.99 \\
        \textbf{FecMap}	& \textbf{83.10}& \textbf{84.92}& \textbf{85.41} \\
\hline\hline
\end{tabular}
\end{footnotesize}
\end{table}
\\\\
% \subsubsection{Parameter Discussion}
4.4.4 \quad Parameter Discussion

\noindent We adjust the number of clients and the size of the dataset to investigate their effects. As can be seen in Fig \ref{Prediction results}(left), the accuracy of FecMap has 2\%-5\% higher than other compared methods when varying the number of clients. The results show that the performance gets better as the number of clients increases. A similar observation could be reached in Fig. \ref{Prediction results}(right) in which the size of the dataset in clients increases. On the other hand, the proposed FecMap delivers the highest accuracy among all the compared state-of-the-art FL models consistently in all experiment cases. Overall, FecMap achieves the best performance on the task of LOP and benefits from more clients and data.
\\\\
% \subsubsection{A Case Study of Online FecMap}
4.4.5 \quad A Case Study of Online FecMap

\noindent In this section, we present a case study on the SE dataset that was collected from the major of software engineering in our institution. The main steps are as follows.

\textbf{STEP 1}: Uploading the input data following the provided example on the homepage of our online system. 

\textbf{STEP 2}: Setting up the training parameters. The parameters include the number of clients, the number of communication rounds, the number of training sessions, the proportion of participants in the training, the learning rate, and the federated learning algorithm. We here set parameters to 5, 20, 10, 0.1, 0.01, and FecMap, respectively.

\textbf{STEP 3}: Training FecMap while monitoring local models' performance. The client models are trained in parallel based on the set number of clients and participation rate. After all participating clients are trained, all local models are uploaded to the server. The server averages all model parameters and distributes the resulting model to each client. The local training loss and accuracy are computed per epoch. 

\textbf{STEP 4}: Calculating and showing the results in terms of various metrics, including prediction accuracy, loss function, and confusion matrix, displayed on the visualization page. 

Fig. \ref{Loss function} shows the loss-function curves and accuracy curves with 20 communication rounds. The FecMap training of this case is completed in 115.557 seconds. Finally, we achieved the following results: the training loss converges in the 17th round; the global model achieves the test accuracy of 66.48\%, while the average accuracy is 68.353\% on the five clients. The obtained test accuracy is lower than the results in Table 2 because this case study used a smaller number of clients.

\section{Conclusion}
 In this paper, we proposed a new FL framework with local subspace learning (LSL) and multi-layer privacy protection (MPP) to deal with data privacy for LOP. LSL is to learn and retain the local features against the global features that are sent into the federated average. MPP is to hierarchically perform and protect the private features, including model-shareable and not-allowed features. Experiment results on our three datasets validate that FecMap has a stable and consistent prediction performance in comparison with other FL models. This study paves the way for data analysis between education institutions with privacy protection.

% \section*{Acknowledgment}
\Acknowledgements{This study was supported in part by the National Natural Science Foundation of China (62272392, U1811262, 61802313), the Key Research and Development Program of China (2020AAA0108500), the Key Research and Development Program of Shaanxi Province (2023-YBGY-405), the fundamental research funds for the central university (D5000230088), and the Higher Research Funding on International Talent Cultivation at NPU (GJGZZD202202).}

\vfill

\end{document}